# The Development and Migration of Concepts from Donor to Borrower Disciplines: Sublanguage Term Use in Hard & Soft Sciences*

Robert M. Losee
School of Information and Library Science
U. of North Carolina
Chapel Hill, NC, 27599-3360

losee@ils.unc.edu

**Abstract**

Academic disciplines, often divided into hard and soft sciences, may be understood as "donor disciplines" if they produce more concepts than they borrow from other disciplines, or "borrower disciplines" if they import more than they originate. Terms used to describe these concepts can be used to distinguish between hard and soft, donor and borrower, as well as individual discipline-specific sublanguages. Using term frequencies, the birth, growth, death, and migration of concepts and their associated terms are examined.

---

*The author wishes to acknowledge valuable discussions with Stephanie Haas, Sydney Pierce, and Lee Ann Roush, as well as the data collection efforts of Brent Stewart

# Introduction

Scholars produce new concepts and related terminology within a disciplinary context. The concepts may spread to others working in the discipline, or even migrate to other disciplines if the concepts are found to have applications beyond the limited set of problems that inspired the original concept. The most useful concepts are those that capture and describe commonly found phenomenon, providing discipline-independent models. These concepts are beneficial to the developer, to scholars in different branches of the same disciplines, and to those in other disciplines. For example, chaotic phenomenon are increasingly studied in a variety of disciplines, and terms such as *chaos* occur with an increased relative frequency in a wide range of academic fields. If we assume that the rise and spread of concepts is reflected in the frequency with which the associated terminology occurs, one may measure the development and spread of concepts by studying the increases and decreases in term occurrences, as well as the relative importance of the concepts. Being able to objectively identify important new developing concepts early in their development may provide a number of benefits, including the ability to support and reward developers of important concepts, as well as to provide support for the further development of concepts that are proving to be important.

Academic disciplines have often been viewed as hard or soft fields of study, with disciplines such as physics or chemistry being viewed as hard sciences, and social sciences such as sociology and political science often being referred to as soft sciences. It is difficult to come up with concrete data and theory that both suggests that there are hard and soft sciences and provides an explanation of the phenomenon. We suggest a classification of disciplines motivated by the desire to explain concept development and migration, into *donor* disciplines, that produce more concepts used by others than use concepts imported from other disciplines, and *borrower* disciplines, that import more concepts than they export. Data presented below supports this donor-borrower distinction, examining concept growth within disciplines, as well as the transfer of specific concepts between disciplines. It is clear from studies of technology development that a majority of equipment is developed in the harder sciences and migrates to the softer sciences (Latour & Woolgar, 1986). Concepts may enter a discipline when scientists themselves move from field to field (Mulkay, 1974) or when scholars who are reasonably competent in two disciplines use concepts and terminology from one discipline to solve problems in another field. Knowing that a field is a donor discipline may signal that the field deserves a disproportionate amount of external funding. Borrower disciplines might receive less external funding than donor disciplines and might want to attract as researchers "borrowers," people knowledgeable about both the borrower discipline and a particular donor field, who would locate material in the donor field that would be of benefit to the borrower field.

We believe that hard sciences are generally donor disciplines and that softer sciences and many professional areas are borrower fields. We provide some preliminary data supporting this notion and describe some characteristics of donor and borrower disciplines and their sublanguages. This is done by developing techniques useful in studying the introduction, development, and death of terms, as well as the migration of terms from one discipline to another. A study of several specific terms suggests that one may qualitatively and quantitatively characterize the migration process, allowing one to study the donating and borrowing processes.

## Academic Disciplines

An academic *discipline* or *field* is a large group of individuals within academia or the professions who are working on a broad range of related research or professional problems. Those within these fields use a *sublanguage,* incorporating the general language of the larger society, as well as a grammar and vocabulary used in a discipline specific manner. For example, in the discipline of library science, the term *classification* has a specific meaning, usually referring to a system used to assign an organizing category to a book, resulting in a label on or near the book's spine and a similar classification number being attached to a bibliographic record describing certain characteristics of the document. A term used in this manner is referred to as being used in a sublanguage sense.

Academic disciplines are usually characterized by the domain of study, for example, physics or sociology are rather clearly defined (except at disciplinary fringes, e.g., physical chemistry, bio-engineering, or social psychology.) These fields may also be characterized by how they relate to other disciplines; for example, some sciences are thought of as "harder" than others. This hierarchy of disciplines has been developed because of a pattern of differences across fields (Cole, 1983; Lodahl & Gordon, 1972; Pierce, 1992; Price, 1986). Researchers in the harder sciences, for example, are more likely to develop testable theories that are useful in predicting future occurrences, quantify their work, achieve consensus within the discipline about the accuracy or usefulness of a theory, and harder sciences grow and make obsolete their older theories faster than do the softer sciences. In addition, researchers in the harder sciences may use terminology more precisely than those in the softer sciences (Mulkay, 1974); we believe that the evidence described below supports this term precision hypothesis.

## Term Frequencies in Sublanguages

The sublanguages used to express the problems and solutions of academic fields differ from each other in many ways, including differing grammatical rules, as well as by the field-specific vocabulary used by specialists (Bonzi, 1984; Damerau, 1990, 1993; Grishman & Kittredge, 1986; Haas, 1995; Haas & He, 1993; Tibbo, 1992). The study of term frequencies and the relationships between frequencies within sublanguages and across sets of sublanguages has proven profitable for studying several problems (Losee & Haas, 1995). A set of terms in a sublanguage may be characterized by the percent of terms used in a discipline specific or technical sense, denoted as $M$, an arbitrary measure. For example, *chaos* is often used in mathematics in the technical sense, implying a specific phenomena where the initial starting point has little to do with the system's state after a moderate period of time, or *chaos* may be be used in a more general sense to mean "disorganized."

Using term frequencies, one study ranks terms by their probability of occurring in a sublanguage with a given or lower frequency, using a set of eight databases compiled by Stephanie Haas (Losee & Haas, 1995). The frequency of term occurrences may be treated as though they are either Poisson or normally distributed across the set of sublanguages. If we consider each document as written in a unique sublanguage with the occurrence of any sublanguage term Poisson distributed, with a different mean in the sublanguage and non-sublanguage documents, it becomes clear that the distribution of a term's frequency in the first or sublanguage document is approximated by the output of a Poisson process (Das Gupta, 1985; Losee, Bookstein, & Yu, 1986; Losee & Haas, 1995; Srinivasan, 1990). When larger numbers of documents exist in a sublanguage, using the normal distribution to model term occurrences may be computationally easier. We refer to this cumulative probability, assuming that it is produced by a Poisson process, as the *Poisson percentile*. Terms that are unique to a particular database have a very high value for the Poisson percentile, approaching 1, while common terms, such as *the*, will have values in a lower range, e.g., less than .4.

After ranking terms by their Poisson percentile, the top ten terms (that were included in the appropriate sublanguage dictionary for the corresponding sublanguage database) were analyzed in (Losee & Haas, 1995) and an average $M$ was computed for these top ten terms, $M^{top}$. The eight disciplines, ranked by their $M^{top}$ values, are: electrical engineering, biology, physics, psychology, mathematics, economics, sociology, and history. A similar value $M^{bottom}$ was computed for the bottom ten terms for each sublanguage. Computing the natural logarithm of the ratio of the average top and bottom $M$ values,

$$M_\Delta = ln(M^{Top}/M^{Bottom}),$$

shows that the positive $M_\Delta$ values are found in what might be considered harder sciences (physics, mathematics, electrical engineering, and biology,) while the softer sciences (sociology, history, economics) have negative values, with psychology having a very small positive value.

These results may be seen as consistent with a term development and migration process. A group of terms having a high Poisson percentile have a high $M$ when they represent concepts that have been developed and retain their technical meaning, and haven't yet migrated to other disciplines. As concepts become more popular, they increasingly become used in a less formal sense, and thus are used in a non-sublanguage sense. A set of terms that develop in the hard sciences (donor disciplines) and are relatively unique to that discipline will have a high Poisson percentile and a high $M$ value, because they haven't moved into more general, non-technical, or less precise usage. They are used in the hard sciences in their intended, field-specific sense.

Terms may have migrated into a discipline from either another domain or from the general language or from both. The lower $M$ values for the most unique terms in the softer sciences suggest that terms in these disciplines are being used in non-specialized senses; this may be due to the "age" of terms and their importation from other disciplines. The term may not have the precise meaning in the new discipline that it had in the donor discipline, and thus will have a lower $M$ value, despite the term being more unique to the sublanguage in question than other terms.

## Term Birth, Migration, and Death in Disciplines

When disciplinary areas such as the hard sciences support research, new concepts and theories are created. These concepts, if beneficial to others in the discipline, experience *intra-disciplinary growth*. If the concepts are general enough, they may benefit other disciplines and *inter-disciplinary* growth and term migration will occur.

We assume that the change in interest in concepts may be tracked by examining the presence or absence of terms in documents. In these donor disciplines, concepts and the associated terms develop and become popular over a period of years. In some cases, this terminology spreads beyond a very small number of authors to become important communicative factors in a discipline. As concepts grow in disciplinary importance, we expect terms associated with the concepts to increase in frequency. Intra-disciplinary growth begins with a growth period that then levels off after several years. When a concept becomes discredited or is replaced, or has been exhaustively studied, a decrease in use will occur (such as happened with the expression *cold fusion*.)

We assume that term growth may be modeled by the diffusion of the concept through a population. Assuming there are $p_m$ potential users of a concept and that $p_t$ have used the concept at time $t$, the rate of increase of term growth may be described as

$$R = \frac{c}{p_m} p_t (p_m - p_t)$$

where $c$ is a constant. For a constant $c = .6$, for example, the growth and spread of the term may be graphed as

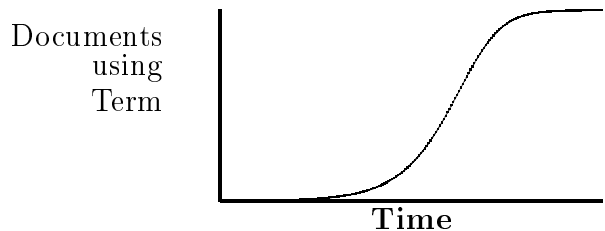

Given term birth, growth, death, and migration, we believe that

- A term's positive growth rate indicates that a concept is being increasingly used in a discipline.

- A term that falls out of favor and is used with decreasing frequency indicates that the associated concept is correspondingly falling out of favor.

- The rate of growth of a term in a discipline is a measure of its importance to the discipline.

- A time lag (measured in years) between the maximum positive growth rate for a term in one discipline and the maximum positive growth rate in another serves as a measure of how fast material is absorbed in the borrowing field from the donating field. It is an indicator of the nature of the channel between the two disciplines, as well as how quickly the receiving field takes in new information.

- The discipline that shows the earliest strong growth peak for a concept may be said to be the donating field.

- Disciplines that show growth peaks after the first peak in the donating discipline, usually with lower growth rates than is found in the donating field, may be said to be borrowing fields.

## Data Analysis

A small number of terms were studied in detail to examine how they increased in frequency within a discipline and as they migrated to other disciplines. An initial set of terms was intellectually extracted from a set of several hundred abstracts obtained from searches of psychological, sociological, and economic bibliographic databases. Searches for terms starting with stems like *mathematic* and *physics* produced abstracts that, in many cases, contained terms from other "harder" sciences. These terms were manually extracted from the abstracts. A smaller list of about 30 terms was developed.

These terms were searched in several CD-ROM databases, covering the psychological, economic, mathematical, educational, and sociological literature. Term frequencies for two year periods were compiled from 1974 to the present for most of the terms. These frequencies were then normalized by dividing the frequency for the two year span by the number of documents in the database published in that two year span. Term frequencies in all cases were based on the binary occurrence of a term in either a document's title or abstract.

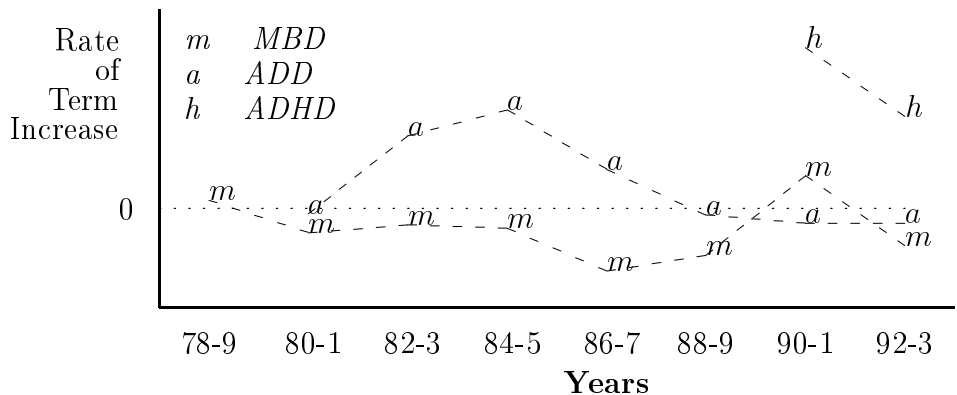

Figure 1: The transition from MBD to ADD to ADHD.

Graphs below show the log rate of increase (or decrease) for terms over a period of time. In these figures, the rate of increase of the frequency of a term's occurrences is positive above the dotted 0 line, while a data point below this line represents a decrease in frequency. The lines are smoothed and those points derived from fewer than 8 documents in a four year period are omitted.

## The Birth, Growth, and Death of Terms

Terms are usually assigned to new concepts during or shortly after the concepts are first described. The terms may be previously unknown or unused terms, such as *quark* or *hypertext*, or concept developers may reuse existing terminology for a new purpose. For example, Claude Shannon was encouraged to use the term *entropy* when describing his communication model because of the commonalities between his formula and the formula describing entropy in classical thermodynamics. Such a term choice may later be regretted due to the confusion it causes; some terms are chosen so that there will be no possible confusion, such as the choice of the terms *charm* and *color* (applied to quarks) in physics.

An interesting snapshot of the death of a concept and its associated term, the rise and fall of its successor, and the rise, in turn, of another successor, is provided by psychology (Figure 1). A specific disorder, or set of symptoms, was referred to in the 1970s as *MBD*, Minimal Brain Dysfunction. The disorder was later referred to as *ADD*, Attention Deficit Disorder in the 1980s, and then began being referred to as *ADHD*, Attention Deficit Hyperactivity Disorder, in the late 1980s. The change from *MBD* to *ADD* reflected a move from focusing on a general set of "soft" neurological signs to a focus on the specific symptoms found in MBD, e.g., ADD. *ADD* was often used with a qualifier, "hyperactive" or "non-hyperactive," with the former being the most common. When the *Diagnostic and Statistical Manual* (DSM) of the American Psychiatric Association began using ADHD as clarifying terminology, its use in psychological literature quickly began to replace the use of *ADD*. Each change represents a move to a more accurate statement of the nature of the syndrome, a perspective that is consistent with the dominant paradigm and with newer scientific data and knowledge.

Note that because the term *ADD* is a common English language term, only documents with *ADD* and the term *attention* were used. A visual examination of several of these documents found them to all be about the diagnostic category.

Figure 1 shows the decreasing rate of occurrences of *MBD* in the early 1980s as the term *ADD* increased in popularity. The rate of increase for *ADD* peaks about 1984, with the rate remaining positive until about 1988, when psychologists began using the term *ADHD*. The rate of increase for *ADHD* peaked about 1990. Note that the greater rate increase for *ADHD* than for *ADD* suggests that the concept underlying ADHD has been more rapidly accepted by the psychological community than was ADD and that ADHD may be considered "more important" or "more useful" than was ADD. Note that the increased rate of occurrence of *MBD* in 1990 appears to represent a noisy fluctuation in a relatively small number of documents.

Another shift in psychological paradigms and nomenclature is currently underway and will allow us to make a prediction that might be verified in the future. The term *MPD*, Multiple Personality Disorder, is being replaced by *DID*, Dissociative Identity Disorder. This reflects a change in the theory underlying the treatment of the disorder from viewing this syndrome as reflecting multiple individual personalities to a view of a single deviant identity. Given the data associated with ADHD, we suspect that a negative growth rate will be seen for *MPD* beginning about the time that a positive growth rate is seen for *DID*. The peak for the positive growth rate for *DID* will probably occur near the fifth year of *DID*'s growth.

## Watching Terms Migrate

Figure 2 shows the rate of growth of the term *chaos*. The concept of chaos has been developing over several decades (Prigogine & Stengers, 1984) but has only began to receive wide acceptance during the past 15 years or so. The term saw a rapid increase in use within mathematics, reaching its maximum growth rate around 1980. The growth continued until recently. The term occurs in the literature of education at an increasing rate, with a weak recent increase in growth rate.

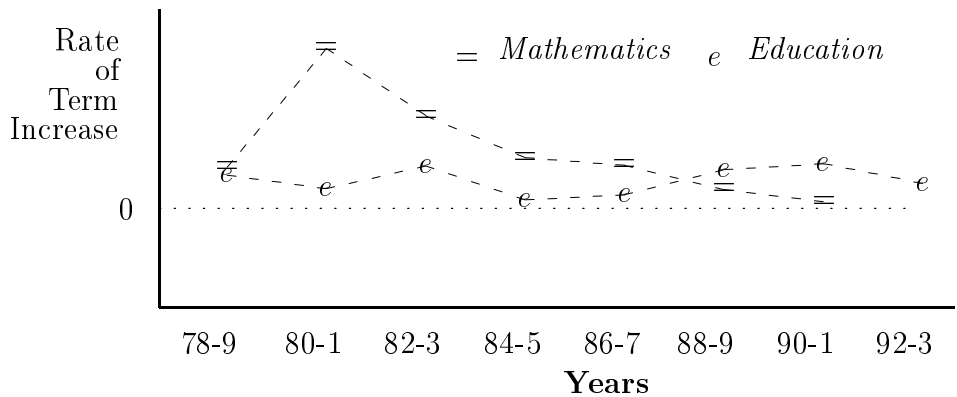

Figure 2: Chaos

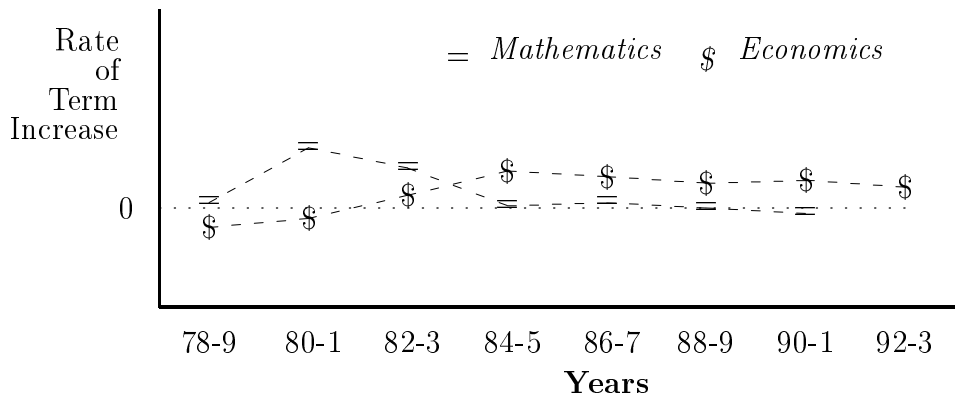

Figure 3: Nonlinear

The growth rate of *chaos* in mathematics is obviously far greater that the growth rate found in educational literature. This is due to its greater relative importance to mathematics. The concept grew rapidly in mathematics, providing a greater body of research which, in turn, helped spread the concept outwards to other fields. The rate of growth is lower in education because it is not as important to that discipline.

The ten year lag between the peak growth periods for the two disciplines is cause by the slow movement from one to the other if we assume that the term and related concepts originated in the first discipline. It also measures the *intellectual distance* between the two fields and the rate at which education absorbed material from mathematics, providing information about how rapidly scholar's in education absorb material from other disciplines. We believe that the time lag between each pair of academic disciplines may be a constant. This will be the object of further study.

Figure 3 shows the change in the rate of use of *nonlinear*, a common term in mathematics that has seen increased numbers of applications recently. Nonlinear systems have become more interesting to researchers, particularly in artificial intelligence related disciplines. The term *nonlinear* increases in frequency in mathematics, with the highest reate of increase occurring about 1980. The term increases at a slower rate in economics, lagging mathematics by about 4 years. The greater rate of increase for *nonlinear* in mathematics than in economics suggests that the concept of nonlinearity and the recent developments in its application are of greater importance to mathematicians than to economists. The lag between the peaks for the two disciplines again suggests that term migration occurred, with economics absorbing at a faster rate than education. A similar phenomena is seen in Figure 4 for the term *convex*, which shows a similar term growth peak lag to what was seen with *nonlinear*.

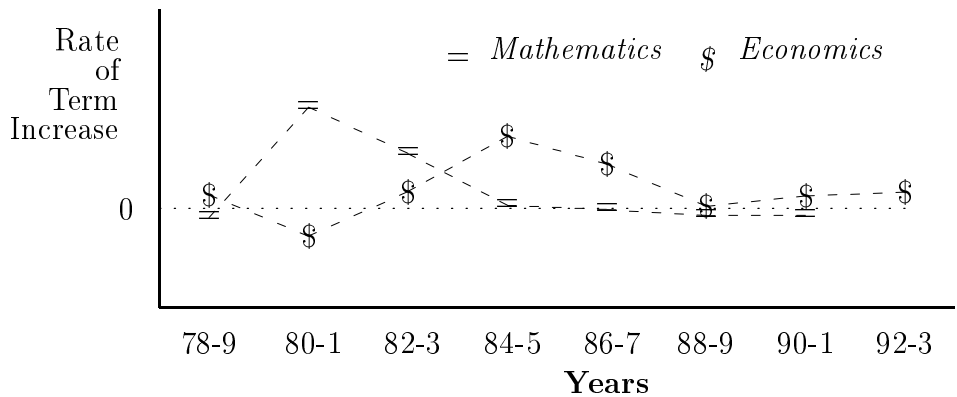

Figure 4: Convex

The lag between mathematics and economics is smaller than the lag between mathematics and education. This is because of the greater intellectual distance between the disciplines.

Further research will be necessary to establish a causal link between the donor and borrower disciplines. This will take the form of citation analysis of the early literature in the borrower field to determine whether, in fact, the donor field, as determined using the above techniques, was the source for the concepts.

# Conclusion

Scientific concepts are developed, sometimes grow, and on rare occasions become useful and popular enough that they migrate to other disciplines. The growth and movement of these concepts can be studied by chronological examination of term frequencies within and across various disciplines. Using these methods, we have provided evidence suggesting that term growth, death, and migration takes place.

Earlier sublanguage research suggested that the terms most unique to a sublanguage were more likely to be used in a sublanguage sense in the harder sciences than in softer sciences (Losee & Haas, 1995). We have suggested here that this phenomena is due to terms developing in the harder donor disciplines and then migrating to other disciplines. Terms that are most unique in the hard, donor disciplines are used in their original, specific sense, while those most unique in the softer, borrowing disciplines are more likely to be used in a general sense, having arrived in the borrowing discipline with a meaning or sense that differs significantly from its original meaning and thus with a less precise meaning than in the originating discipline. This provides both data and an explanation as to why some sciences appear to be different from others. For example, unlike studies that show different perceptions of hard sciences and soft sciences, our data suggests both a measure of a factor like "hardness" as well as provides an explanation of the underlying phenomenon that produces "hardness" (donation) and "softness" (borrowing).

Terms such as *chaos, nonlinear,* and *convex* grew in frequency in the literature of mathematics, considered here to have been the originating discipline. These terms later increased in popularity in other disciplines as the concepts migrated to these other disciplines.

If we accept the hypothesis that the rate of term growth is indicative of the importance of the concept to the discipline, we may conclude that these terms are more important to the originating discipline than to the borrowing discipline, as one would expect. We have shown that for several sublanguage terms, the growth rate was higher for the originating discipline than for the borrowing discipline.

While there are other measures of research importance, such as citation counts, the rate of growth may prove useful as another general indicator of concept importance. Using the growth rate of a term as an indicator of research importance avoids the problem of trying to normalize the measure for the size of the user population, as may be desirable in citation counting. Using the term growth rate also makes it unnecessary to determine what the unit of research should be and would have the effect of decreasing the pressure to cite based on what may be "political" considerations. What matters to a science is how useful others find the research and the concepts developed, which, in many instances, may be measured by examining the increasing (or decreasing) rate of term use.

In addition, the objective identification of important concepts early in their development can lead to financial support to further the development of the innovative work. Identifying donor disciplines may lead to further support for these crucial fields, while the identification of borrowing fields, along with their primary donor fields, may encourage the development of specialists, "borrowers," who are capable of understanding the developments in the donor fields that can be of use to their borrower field.

# References

Bonzi, S. (1984). Terminological consistency in abstract and concrete disciplines. *Journal of Documentation*, *40*(4), 247–263.

Cole, S. (1983). The hierarchy of the sciences?. *American Journal of Sociology*, *89*(1), 111–139.

Damerau, F. J. (1990). Evaluating computer-generated domain-oriented vocabularies. *Information Processing and Management*, *26*(6), 791–801.

Damerau, F. J. (1993). Generating and evaluating domain-oriented multi-word terms from texts. *Information Processing and Management*, *29*(4), 433–447.

Das Gupta, P. (1985). *An Investigation into the Two-Poisson Model of Automatic Indexing*. Ph.D. thesis, Syracuse University.

Grishman, R., & Kittredge, R. (Eds.). (1986). *Analyzing Language in Restricted Domains*. Lawrence Erlbaum Associates, Hillsdale, NJ.

Haas, S. W. (1995). Domain terminology patterns in different disciplines: Evidence from abstracts. In *Proceedings of the Fourth Annual Symposium on Document Analysis and Information Retrieval*, pp. 137–146 Las Vegas, NV.

Haas, S. W., & He, S. (1993). Toward the automatic identification of sublanguage vocabulary. *Information Processing and Management*, *29*(6), 721–732.

Latour, B., & Woolgar, S. (1986). *Laboratory Life: The Construction of Scientific Facts*. Princeton Univ. Press, Princeton, NJ.

Lodahl, J. B., & Gordon, G. (1972). The structure of scientific fields and the functioning of university graduate departments. *American Sociological Review*, *37*, 57–72.

Losee, R. M., Bookstein, A., & Yu, C. T. (1986). Probabilistic models for document retrieval: A comparison of performance on experimental and synthetic databases. In *ACM Conference on Research and Development in Information Retrieval*, pp. 258–264.

Losee, R. M., & Haas, S. W. (1995). Sublanguage terms: Dictionaries, usage, and automatic classification. *Journal of the American Society for Information Science*, *46*(7), 519–529.

Mulkay, M. (1974). Conceptual displacements and migration in science. *Science Studies*, *4*, 205–234.

Pierce, S. J. (1992). On the origin and meaning of bibliometric indicators: Journals in the social sciences, 1886–1985. *Journal of the American Society for Information Science, 43*(7), 477–487.

Price, D. (1986). *Little Science, Big Science ...and Beyond.* Columbia University Press, New York.

Prigogine, I., & Stengers, I. (1984). *Order Out of Chaos.* Bantam.

Srinivasan, P. (1990). On generalizing the two-Poisson model. *Journal of the American Society for Information Science, 41*(1), 61–66.

Tibbo, H. R. (1992). Abstracting across the disciplines: A content analysis of abstracts from the natural sciences, the social sciences, and the humanities with implications for standardization and online information retrieval. *Library and Information Science Research, 14*, 31–56.